\documentclass{article} 
\usepackage{iclr2016_conference,times}
\usepackage{hyperref}
\usepackage{url}

\usepackage{multicol}
\usepackage{multirow}
\usepackage{booktabs}
\usepackage{graphicx}
\graphicspath{{../diagrams/},{./}}
\usepackage{subcaption}
\usepackage{amssymb}
\usepackage{amsmath}
\usepackage{bm}
\usepackage{color}
\usepackage{calc}
\usepackage{tikz}
\usetikzlibrary{bayesnet}

\DeclareMathOperator{\Tr}{\mathrm{Tr}}

\iclrfinalcopy

\title{Recurrent Gaussian Processes}

\author{C\'esar Lincoln C. Mattos$^1$, Zhenwen Dai$^2$, Andreas Damianou$^3$, Jeremy Forth$^4$, \\
\textbf{Guilherme A. Barreto$^5$ \& Neil D. Lawrence$^6$} \\
$^{1,5}$Federal University of Cear\'a, Fortaleza, Cear\'a, Brazil \\
$^{2,3,6}$University of Sheffield, Sheffield, UK \\
$^1$\texttt{cesarlincoln@terra.com.br} \\ $^{2,3}$\texttt{\{z.dai,andreas.damianou\}@sheffield.ac.uk} \\
$^4$\texttt{jforth@iweng.org} \\
$^5$\texttt{gbarreto@ufc.br} \\
$^6$\texttt{N.Lawrence@dcs.sheffield.ac.uk}
}

\begin{document}

\maketitle

\begin{abstract}
We define Recurrent Gaussian Processes (RGP) models, a general family of Bayesian nonparametric models with recurrent GP priors which are able to learn dynamical patterns from sequential data. Similar to Recurrent Neural Networks (RNNs), RGPs can have different formulations for their internal states, distinct inference methods and be extended with deep structures. In such context, we propose a novel deep RGP model whose autoregressive states are latent, thereby performing representation and dynamical learning simultaneously. To fully exploit the Bayesian nature of the RGP model we develop the Recurrent Variational Bayes (REVARB) framework, which enables efficient inference and strong regularization through coherent propagation of uncertainty across the RGP layers and states. We also introduce a RGP extension where variational parameters are greatly reduced by being reparametrized through RNN-based sequential recognition models. We apply our model to the tasks of nonlinear system identification and human motion modeling. The promising obtained results indicate that our RGP model maintains its highly flexibility while being able to avoid overfitting and being applicable even when larger datasets are not available.
\end{abstract}


\section{Introduction}

The task of learning patterns from sequences is an ongoing challenge for the machine learning community. Recurrent models are able to learn temporal patterns by creating internal memory representations of the data dynamics. A general recurrent model, comprised of external inputs $\bm{u}_i$, observed outputs $\bm{y}_i$ and hidden states $\bm{x}_i$, is given by
\begin{align}
\label{EQ:REC_1}
\bm{x}_i & = f(\bm{x}_{i-1}, \bm{u}_{i-1}) + \bm{\epsilon}_i^x, \\
\label{EQ:REC_2}
\bm{y}_i & = g(\bm{x}_{i}) + \bm{\epsilon}_i^y,
\end{align}
where $i$ is the instant of observation, $f(\cdot)$ and $g(\cdot)$ are unknown nonlinear functions respectively called \textit{transition} and \textit{observation} functions, $ \bm{\epsilon}_i^x \sim \mathcal{N}(\bm{\epsilon}_i^x | \bm{0}, \sigma_x^2 \bm{I})$ and $\bm{\epsilon}_i^y \sim \mathcal{N}(\bm{\epsilon}_i^y | \bm{0}, \sigma_y^2 \bm{I})$ are respectively Gaussian transition and observation noises, and $\bm{I}$ is the identity matrix. The recurrent nature of the model is expressed by the state variables $\bm{x}_i$, which are dependent on their past values, allowing past patterns to have influence in future outputs. 

In recurrent parametric models, such as Recurrent Neural Networks (RNN), both the transition and observation functions are modeled with weight matrices $\bm{W}$, $\bm{U}$, $\bm{V}$ and nonlinear element-wise activation functions $\phi_f(\cdot)$ and $\phi_g(\cdot)$:
\begin{align}
\label{EQ:RNN_1}
\bm{x}_i & = \phi_f(\bm{W}^\top\bm{x}_{i-1}, \bm{U}^\top\bm{u}_{i-1}), \\
\label{EQ:RNN_2}
\bm{y}_i & = \phi_g(\bm{V}^\top\bm{x}_{i}).
\end{align}

As argued by \citet{pascanu2013}, the basic recurrent structure in Eq. \ref{EQ:RNN_1} can be made \textit{deep}, for example by adding multiple hidden layers comprised of multiple transition functions, where the output of each layer is used as the input of the next one.

The difficulties related to learning dynamical structures from data \citep{bengio1994} have motivated the proposal of several RNN architectures in the literature, such as time-delay neural networks \citep{lang1990}, hierarchical RNNs \citep{el1995}, nonlinear autoregressive with exogenous inputs (NARX) neural networks \citep{lin1996}, long short-term memory networks \citep{hochreiter1997}, deep RNNs \citep{pascanu2013} and the RNN encoder-decoder \citep{cho2014}. The usefulness of RNNs has been demonstrated in interesting applications, such as music generation \citep{boulanger2012}, handwriting synthesis, \citep{graves2013a} speech recognition \citep{graves2013b}, and machine translation \citep{cho2014}.

However, one well known limitation of parametric models, such as RNNs, is that they usually require large training datasets to avoid overfitting and generalization degradation. In contrast Bayesian nonparametric methods, such as Gaussian Processes (GP) models, often perform well with smaller datasets. In particular GP-based models are able to propagate uncertainty through their different structural components, something which ensures that when data is not present in a particular region of input space the predictions do not become over confident.

The general recurrent Eqs. \ref{EQ:REC_1} and \ref{EQ:REC_2} have been widely studied in the control and dynamical system identification community as either non-linear auto-regressive models with exogenous inputs (NARX) models  or state-space models (SSM). Here we are particularly interested in the Bayesian approach to those models \citep{peterka1981}. In this context, several GP-NARX models have been proposed in the literature \citep{murray1999,solak2003,kocijan2005}. However, these models do not propagate the past states' uncertainty through the transition function during the training or prediction phase. \cite{girard2003,Damianou:semidescribed15} rectify this problem. Nevertheless, in all of the above standard NARX approaches the autoregressive structure is performed directly with the observed outputs, which are noisy.

A more general alternative to standard NARX models is the use of SSMs. Such structures have been explored recently by the GP community. \cite{frigola2014} proposed a variational GP-SSM where both the transition and observation functions can have GP priors. Although they present results exclusively for the case where only the transition is modeled by a GP, while the observation has a parametric form. Conversely, Moreover, the inference required an additional smoothing step with, for example, a sequential Monte Carlo algorithm. \cite{svensson2015} also consider a GP-SSM, but with a reduced-rank structure, and perform inference following a fully Bayesian approach, using a particle MCMC algorithm.

All the aforementioned dynamic GP models contain recurrent structures. Each model makes a particular choice for the definition of the states $\bm{x}_i$ and the algorithm used to perform inference. 
Because all these GP models incorporate recurrent structures we refer to this general class of models as the \emph{Recurrent GP (RGP)} family of methods. These are models such as in Eqs \eqref{EQ:REC_1} and \eqref{EQ:REC_2} which incorporate GP priors for the transition and/or observation functions. Inspired by developments in the RNN community we propose a novel RGP model which introduces \emph{latent autoregression} and is embedded in a new variational inference procedure named \emph{Recurrent Variational Bayes (REVARB)}. Our formulation aims to tackle some issues of past RGP structures. Our algorithm allows the RGP class of models to easily be extended to have deep structures, similar to deep RNNs. Furthermore, we develop an extension which combines the RGP and RNN technologies by reparameterizing the means of REVARB's variational distributions through a new RNN-based recognition model. This idea results in simpler optimization and faster inference in larger datasets.

Recently, \cite{sohl2015} have detailed interesting similarities between the log-likelihood training of RNNs and the variational Bayes training objective in the context of generative models. In the present work we also follow a variational approach with the proposed REVARB framework, but with respect to RGP models.

The rest of the paper is structured as follows. In Section \ref{SEC:GP} we briefly summarize the standard GP for regression. In Section \ref{SEC:RGP} we define the structure of our proposed RGP model. In Section \ref{SEC:REVARB} we describe the REVARB inference method. In Section \ref{SEC:EXPERIMENTS} we present some experiments with REVARB in some challenging applications. We conclude the paper with hints to further work in Section \ref{SEC:CONCLUSIONS}.

\section{Standard GP Model for Regression}
\label{SEC:GP}

In the GP framework, a multiple input single output nonlinear function $f(\cdot)$ applied to a collection of $N$ examples of $D$-dimensional inputs $\bm{X} \in \mathbb{R}^{N \times D}$ is given a multivariate Gaussian prior:
\begin{equation}
\label{EQ:GP}
\bm{f} = f(\bm{X}) \sim \mathcal{N}( \bm{f} | \bm{0}, \bm{K} ),
\end{equation}
where a zero mean vector was considered, $\bm{f} \in \mathbb{R}^N$ and $\bm{K} \in \mathbb{R}^{N \times N}, K_{ij} = k(\bm{x}_i, \bm{x}_j)$, is the covariance matrix, obtained with a covariance (or \textit{kernel}) function $k(\cdot, \cdot)$, which must generate a semidefinite positive matrix $\bm{K}$, for example the exponentiated quadratic kernel:
\begin{equation}
\label{EQ:COV}
k(\bm{x}_i, \bm{x}_j) = \sigma_f^2 \exp \left[ -\frac{1}{2} \sum_{d=1}^{D} w_d^2 (x_{id} - x_{jd})^2 \right],
\end{equation}
where the vector $\bm{\theta} = [\sigma_f^2, w_1^2, \dots, w_D^2]^\top$ is comprised of the hyperparameters which characterize the covariance of the model.

If we consider a Gaussian likelihood $p(\bm{y} | \bm{f}) = \mathcal{N}(\bm{y} | \bm{f}, \sigma_y^2 \bm{I})$ relating the observations $\bm{y}$ and the unknown values $\bm{f}$, inference for a new output $f_*$, given a new input $\bm{x}_*$, is calculated by:
\begin{equation}
\label{EQ:GAUSSIAN_INFERENCE}
p(f_*|\bm{y},\bm{X},\bm{x}_*) = \mathcal{N}\left( f_* | \bm{k}_{*N} (\bm{K} + \sigma_y^2\bm{I})^{-1} \bm{y}, k_{**} - \bm{k}_{*N} (\bm{K} + \sigma_y^2\bm{I})^{-1} \bm{k}_{N*} \right),
\end{equation}
where $\bm{k}_{*N} = [k(\bm{x}_*,\bm{x}_1), \cdots, k(\bm{x}_*,\bm{x}_N)]$, $\bm{k}_{N*} = \bm{k}_{*N}^\top$ and $k_{**} = k(\bm{x}_*,\bm{x}_*)$. The predictive distribution of $y_*$ is similar to the one in Eq. (\ref{EQ:GAUSSIAN_INFERENCE}), but the variance is added by $\sigma_y^2$.

The vector of hyperparameters $\bm{\theta}$ can be extended to include the noise variance $\sigma_y^2$ and be determined with the maximization of the marginal log-likelihood $\log p(\bm{y} | \bm{X}, \bm{\theta})$ of the observed data, the so-called \textit{evidence} of the model. The optimization is guided by the gradients of the evidence with respect to each component of the vector $\bm{\theta}$. It is worth mentioning that such optimization can be seen as the model selection step of obtaining a plausible GP model from the training data.

\section{Our Recurrent GP Model}
\label{SEC:RGP}

We follow an alternative SSM approach where the states have an autoregressive structure. Differently from standard NARX models, the autoregression in our model is performed with \textit{latent} (non-observed) variables. Thus, given $L$ lag steps and introducing the notation $\bm{\bar{x}}_i = [x_i, \cdots, x_{i-L+1}]^\top$ we have
\begin{align}
\label{EQ:LAGP_1}
x_i & = f( \bm{\bar{x}}_{i-1}, \bm{\bar{u}}_{i-1}) + \epsilon_i^x, \\
\label{EQ:LAGP_2}
\bm{y}_i & = g(\bm{\bar{x}}_i) + \bm{\epsilon}_i^y,
\end{align}
where $\bm{\bar{u}}_{i-1} = [u_{i-1}, \cdots, u_{i-L_u}]^\top$ and $L_u$ is the number of past inputs used. Even if the output of the transition function in Eq. \ref{EQ:LAGP_1} is chosen to be 1-dimensional, it should be noticed that the actual hidden state $\bm{\bar{x}}_i \in \mathbb{R}^L$ is multidimensional for $L > 1$.

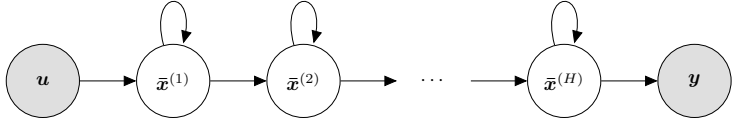
\begin{figure}[!ht]
\centering
\newcommand{\nodeSize}{1.3cm}

\resizebox{0.7\linewidth}{!}{
\begin{tikzpicture}

%

  \node[obs, minimum size=\nodeSize] (y) {$\bm{y}$};
  \node[latent, left=of y, minimum size=\nodeSize] (xh) {$\bm{\bar{x}}^{(H)}$};
  \node[left=of xh, minimum size=\nodeSize]  (dots) {$\cdots$};
  \node[latent, left=of dots, minimum size=\nodeSize] (x2) {$\bm{\bar{x}}^{(2)}$};
  \node[latent, left=of x2, minimum size=\nodeSize] (x1) {$\bm{\bar{x}}^{(1)}$};
  \node[obs, left=of x1, minimum size=\nodeSize] (u) {$\bm{u}$};
  
  \edge {u}{x1};
  \edge {x1}{x2};
  \edge[loop above] {x1}{x1};
  \edge[loop above] {x2}{x2};
  \edge {x2}{dots};  
  \edge {dots}{xh}; 
  \edge[loop above] {xh}{xh};
  \edge {xh}{y}; 

\end{tikzpicture}
}
\caption{RGP graphical model with $H$ hidden layers.}
\label{FIG:DIA_RGP}
\end{figure}

If we have $H$ transition functions, each one comprising a hidden layer, it naturally gives rise to the deep structure
\begin{align}
\label{EQ:RGP1}
x_i^{(h)} & = f^{(h)}\left( \bm{\hat{x}}_i^{(h)} \right) + \epsilon_i^{(h)}, & \bm{f}^{(h)} & \sim \mathcal{N}\left(\bm{0},\bm{K}_f^{(h)}\right), & \quad 1 \leq h \leq H \\
\label{EQ:RGP2}
\bm{y}_i & = f^{(H+1)}\left(\bm{\hat{x}}_i^{(H+1)}\right) + \epsilon_i^{(H+1)}, & \bm{f}^{(H+1)} & \sim \mathcal{N}\left(\bm{0},\bm{K}_f^{(H+1)}\right) &
\end{align}
where we put GP priors with zero mean and covariance matrix $\bm{K}_f^{(h)}$ on the unknown functions $f(\cdot)^{(h)}$, the noise in each layer is defined as $\epsilon_i^{(h)} \sim \mathcal{N}( 0, \sigma_h^2 )$ and the upper index differentiates variables and functions from distinct layers. We also introduce the notation
\begin{equation}
\label{EQ:X_HAT}
\bm{\hat{x}}_i^{(h)} =
\begin{cases}
\left[\bm{\bar{x}}_{i-1}^{(1)}, \bm{\bar{u}}_{i-1}\right]^\top = \left[ \left[x_{i-1}^{(1)}, \cdots, x_{i-L}^{(1)}\right], [u_{i-1}, \cdots, u_{i-L_u}] \right]^\top , & \mathrm{if} \; h=1,  \\
\left[\bm{\bar{x}}_{i-1}^{(h)}, \bm{\bar{x}}_i^{(h-1)}\right]^\top = \left[ \left[x_{i-1}^{(h)}, \cdots, x_{i-L}^{(h)}\right], \left[x_i^{(h-1)}, \cdots, x_{i-L+1}^{(h-1)}\right] \right]^\top , & \mathrm{if} \; 1 < h \leq H,  \\
\bm{\bar{x}}_i^{(H)} = \left[x_i^{(H)}, \cdots, x_{i-L+1}^{(H)}\right]^\top , & \mathrm{if} \; h=H+1.
\end{cases}
\end{equation}
The graphical model for the RGP is presented in Fig. \ref{FIG:DIA_RGP}, where we kept the general states $\bm{\bar{x}}^{(h)}$ to make the recurrent connections more clear. It should be noted that the standard GP-NARX and GP-SSM are also RGPs, but with different states structure.

Our RGP model, as defined by Eqs. \ref{EQ:RGP1} and \ref{EQ:RGP2}, can be seen as a special case of the Deep GP framework \citep{damianou2013,damianou:thesis15} where the priors of the latent variables in each hidden layer follow the autoregressive structure of Eq. \ref{EQ:X_HAT}. 


We emphasize that our model preserves the non-observed states of standard SSMs but avoids the ambiguities of generic multidimensional states by imposing a latent autoregressive structure. In the next section, we explain how this novel RGP model can be trained using the REVARB framework.

\section{Recurrent Variational Bayes (REVARB)}
\label{SEC:REVARB}

Inference is intractable in our RGP model because we are not able to get analytical forms for the posterior of $\bm{f}^{(h)}$ or the marginal likelihood of $\bm{y}$. In order to tackle such intractabilities, we apply a novel variational approximation scheme named REVARB.

REVARB is based on the variational sparse framework proposed by \citet{titsias2009}, thus, we start by including to each layer $h$ a number of $M$ inducing points $\bm{z}^{(h)} \in \mathbb{R}^M$ evaluated in $M$ pseudo-inputs $\bm{\zeta}^{(h)} \in \mathbb{R}^D$ such as that $\bm{z}^{(h)}$ are extra samples of the GP that models $f^{(h)}(\cdot)$ and $p\left(\bm{z}^{(h)}\right) = \mathcal{N}\left(\bm{z}^{(h)}\middle|\bm{0},\bm{K}_z^{(h)}\right)$, where $\bm{K}_z^{(h)}$ is the covariance matrix obtained from $\bm{\zeta}^{(h)}$. Considering a model with $H$ hidden layers and 1-dimensional outputs, the joint distribution of all the variables is given by:
\begin{align}
\begin{split}
\label{EQ:JOINT}
& p\left(\bm{y}, \bm{f}^{(H+1)}, \bm{z}^{(H+1)}, \left\{\bm{x}^{(h)}, \bm{f}^{(h)}, \bm{z}^{(h)}\right\}\middle|_{h=1}^{H} \right) = \\
& \left( \prod_{i=L+1}^N p\left(y_i \middle| f_i^{(H+1)}\right) p\left(f_i^{(H+1)} \middle| \bm{z}^{(H+1)}, \bm{\hat{x}}_i^{(H)}\right) \prod_{h=1}^{H} p\left(x_i^{(h)} \middle| f_i^{(h)}\right) p\left(f_i^{(h)} \middle| \bm{z}^{(h)}, \bm{\hat{x}}_i^{(h)}\right) \right) \\
& \left( \prod_{h=1}^{H+1} p\left(\bm{z}^{(h)}\right) \right) \left( \prod_{i=1}^{L} \prod_{h=1}^{H} p\left(x_i^{(h)}\right) \right).
\end{split}
\end{align}

By applying Jensen's inequality, similar to the standard variational approach, we can obtain a lower bound to the log-marginal likelihood $\log p(\bm{y})$ \citep{bishop2006}:
\begin{equation}
\label{EQ:BOUND1}
\log p(\bm{y}) \geq \int_{\bm{f},\bm{x},\bm{z}} Q \log \left[ \frac{p\left(\bm{y}, \bm{f}^{(H+1)}, \bm{z}^{(H+1)}, \left\{\bm{x}^{(h)}, \bm{f}^{(h)}, \bm{z}^{(h)}\right\}\middle|_{h=1}^{H} \right)}{Q} \right],
\end{equation}
where $Q$ is the variational distribution. We choose the following factorized expression:
\begin{equation}
\label{EQ:VARIATIONAL_Q}
Q = \left( \prod_{h=1}^{H} q\left(\bm{x}^{(h)}\right) \right) \left( \prod_{h=1}^{H+1} q\left(\bm{z}^{(h)}\right) \right) \left( \prod_{i=L+1}^N \prod_{h=1}^{H+1} p\left(f_i^{(h)} \middle| \bm{z}^{(h)}, \bm{\hat{x}}_i^{(h)}\right) \right).
\end{equation}

Considering a mean-field approximation, each term is given by
\begin{align}
\label{EQ:QX}
q\left(\bm{x}^{(h)}\right) & = \prod_{i=1}^N \mathcal{N}\left(x_i^{(h)} \middle| \mu_i^{(h)},\lambda_i^{(h)}\right), \\
q\left(\bm{z}^{(h)}\right) & = \mathcal{N}\left( \bm{z}^{(h)} \middle| \bm{m}^{(h)}, \bm{\Sigma}^{(h)}\right), \\
p\left(f_i^{(h)} \middle| \bm{z}^{(h)}, \bm{\hat{x}}_i^{(h)}\right) & = \mathcal{N}\left(f_i^{(h)} \middle| \left[\bm{a}_f^{(h)}\right]_i, \left[\bm{\Sigma}_f^{(h)}\right]_{ii}\right),
\end{align}
\begin{equation*}
\text{where} \; \; \bm{a}_f^{(h)} = \bm{K}_{fz}^{(h)} \left(\bm{K}_z^{(h)}\right)^{-1} \bm{z}^{(h)} \text{\; \; and \; \;}
\bm{\Sigma}_f^{(h)} = \bm{K}_f^{(h)} - \bm{K}_{fz}^{(h)} \left(\bm{K}_z^{(h)}\right)^{-1} \left(\bm{K}_{fz}^{(h)}\right)^\top.
\end{equation*}
In the above, $\mu_i^{(h)}$, $\lambda_i^{(h)}$, $\bm{m}^{(h)}$ and $\bm{\Sigma}^{(h)}$ are variational parameters, $\bm{K}_f^{(h)}$ is the standard kernel matrix obtained from $\bm{\hat{x}}^{(h)}$, $\bm{K}_z^{(h)}$ is the sparse kernel matrix calculated from the pseudo-inputs $\bm{\zeta}^{(h)}$ and $\bm{K}_{fz}^{(h)} = k(\bm{\hat{x}}^{(h)},\bm{\zeta}^{(h)}) \in \mathbb{R}^{N \times M}$.

The variational distribution in Eq. \ref{EQ:QX} indicates that the latent variables $\bm{x}^{(h)}$ are related to $2N$ variational parameters. In standard variational GP-SSM, such as in \cite{frigola2014} we would have a total of $2ND$ parameters, for $D$-dimensional states, even for a diagonal covariance matrix in the posterior. Such reduction of parameters in the mean-field approximation was enabled by the latent autoregressive structure of our model.

Replacing the variational distribution in the Eq. \ref{EQ:BOUND1} and working the expressions we are able to optimally eliminate the variational parameters $\bm{m}^{(h)}$ and $\bm{\Sigma}^{(h)}$, obtaining the final form of the lower bound, presented in the included appendix. We have to compute some statistics that come up in the full bound:
\begin{equation}
\label{EQ:PSI_STATS}
\begin{array}{@{}l@{}}
\Psi_0^{(h)} = \Tr\left(\left\langle \bm{K}_f^{(h)} \right\rangle_{q(\cdot)^{(h)}}\right) \\
\bm{\Psi}_1^{(h)} = \left\langle \bm{K}_{fz}^{(h)} \right\rangle_{q(\cdot)^{(h)}} \\
\bm{\Psi}_2^{(h)} = \left\langle \left(\bm{K}_{fz}^{(h)}\right)^\top \bm{K}_{fz}^{(h)} \right\rangle_{q(\cdot)^{(h)}}
\end{array} \Rightarrow q(\cdot)^{(h)} =
\begin{cases}
q\left(\bm{x}^{(1)}\right), & \mathrm{if} \; h=1,  \\
q\left(\bm{x}^{(h)}\right)q\left(\bm{x}^{(h-1)}\right), & \mathrm{if} \; 1 < h \leq H,  \\
q\left(\bm{x}^{(H)}\right), & \mathrm{if} \; h=H+1,
\end{cases}
\end{equation}
where $\langle \cdot \rangle_{q(\bm{x}^{(h)})}$ means expectation with respect to the distribution $q\left(\bm{x}^{(h)}\right)$, which itself depends only on the variational parameters $\mu_i^{(h)}$ and $\lambda_i^{(h)}$. All the expectations are tractable for our choice of the exponentiated quadratic covariance function and follow the same expressions presented by \cite{titsias2010}. The bound can be optimized with the help of analytical gradients with respect to the kernel and variational hyperparameters.

The REVARB framework allows for a natural way to approximately propagate the uncertainty during both training and prediction. For testing, given a new sequence of external inputs, we can calculate the moments of the predictive distribution of each layer by recursively applying the results introduced in \cite{girard2003}, with predictive equations presented in the included appendix.

\subsection{Sequential RNN-based recognition model}
From Eq. \eqref{EQ:QX} it is obvious that the number of variational parameters in REVARB grows linearly with the number of output samples. This renders optimization challenging in large $N$ scenarios. To alleviate this problem we propose to constrain the variational means $\left\{ \mu^{(h)}_i \right\}, \forall h,i$ using RNNs. More specifically, we have:
\begin{equation}
\mu_i^{(h)} = g^{(h)}\left( \bm{\hat{x}}_{i-1}^{(h)} \right), \text{where}\; g (\bm{x})  = \bm{V}_{L_N}^\top \boldsymbol \phi_{L_N}(\bm{W}_{L_N-1} \boldsymbol \phi_{L_N-1}(\cdots \bm{W}_2 \boldsymbol \phi_1(\bm{U}_1\bm{x}))),
\end{equation}
$\bm{W},\bm{U}$ and $\bm{V}$ are parameter matrices, $\boldsymbol \phi(\cdot)$ denotes the hyperbolic tangent activation function and $L_N$ denotes the depth of the neural network. We refer to this RNN-based constraint as the \emph{sequential recognition model}. Such model directly captures the transition between the latent representation across time. This provides a constraint over the variational posterior distribution of the RGP that emphasizes free simulation. The recognition model's influence is combined with that of the analytic lower bound in the same objective optimization function. In this way, we no longer need to optimize the variational means but, instead, only the set of RNN weights, whose number does not increase linearly with $N$. Importantly, this framework also allows us to kick-off optimization by random initialization of the RNN weights, as opposed to more elaborate initialization schemes. 
 The recognition model idea relates to the work of \citep{KingmaWelling2013, RezendeEtAl2014}. In our case, however, the recognition model is sequential to agree with the latent structure and its purpose is distinct, because it acts as a constraint in an already analytic variational lower bound. Furthermore, our sequential recognition model acts upon a nonparametric Bayesian model.

\section{Experiments}
\label{SEC:EXPERIMENTS}

In this section we evaluate the performance of our RGP model in the tasks of nonlinear system identification and human motion modeling.

\subsection{Nonlinear System Identification}

We use one artificial benchmark, presented by \cite{narendra1996}, and two real datasets. The first real dataset, named \textit{Actuator} and described by \cite{sjoberg1995} \footnote{Available in the DaISy repository at \url{http://www.iau.dtu.dk/nnbook/systems.html}.}, consists of a hydraulic actuator that controls a robot arm, where the input is the size of the actuator's valve opening and the output is its oil pressure. The second dataset, named \textit{Drives} and introduced by \cite{wigren2010}, is comprised by a system with two electric motors that drive a pulley using a flexible belt. The input is the sum of voltages applied to the motors and the output is the speed of the belt.

In the case of the artificial dataset we choose $L = L_u = 5$ and generate 300 samples for training and 300 samples for testing, using the same inputs described by \cite{narendra1996}. For the real datasets we use $L = L_u = 10$ and apply the first half of the data for training and the second one for testing. The evaluation is done by calculating the root mean squared error (RMSE) of the free simulation on the test data. We emphasize that the predictions are made based only on the test inputs and past predictions.

We compare our RGP model with 2 hidden layers, REVARB inference and 100 inducing inputs with two models commonly applied to system identification tasks: standard GP-NARX and MLP-NARX. We use the MLP implementation from the MATLAB Neural Network Toolbox with 1 hidden layer. We also include experiments with the LSTM network, although the task itself probably does not require long term dependences. The original LSTM architecture by \citet{hochreiter1997} was chosen, with a network depth of 1 to 3 layers and the number of cells at each layer selected to be up to 2048. LSTM memory length was unlimited, and sequence length was chosen initially to be a multiple of the longest duration memory present in the data generative process, and reduced gradually. During experiments with varying LSTM network configurations, it became clear that it was possible in most cases to obtain convergence on the training sets, using a carefully chosen network model size and hyperparameters.  Training was organized around batches, and achieved using a learning rate selected to fall slightly below loop instability, and it was incrementally reduced when instability re-appeared. A batch in this context is the concatenation of fixed length sub-sequences of the temporal data set. Neither gradient limits nor momentum were used.

\begin{figure}[!ht]
\centering
\begin{minipage}{\linewidth}
\begin{subfigure}[b]{.32\linewidth}
\includegraphics[width=\linewidth]{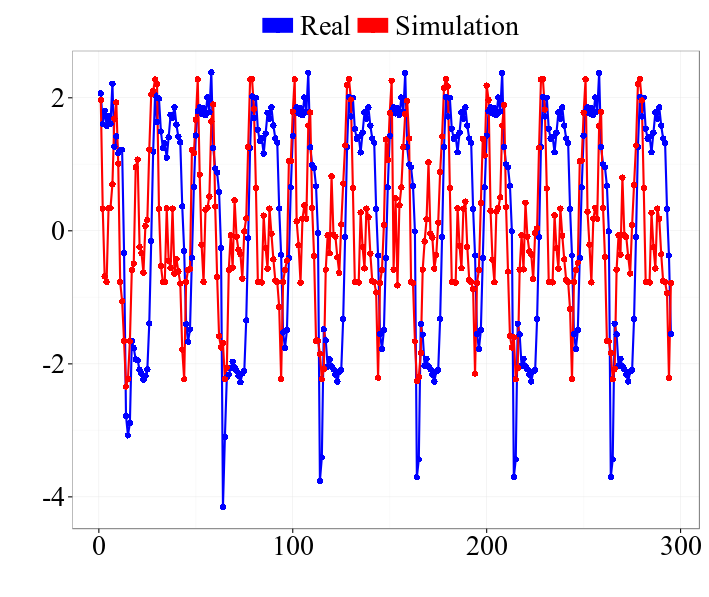}
\caption{MLP-NARX - Artificial dataset.}
\end{subfigure}
\begin{subfigure}[b]{.32\linewidth}
\includegraphics[width=\linewidth]{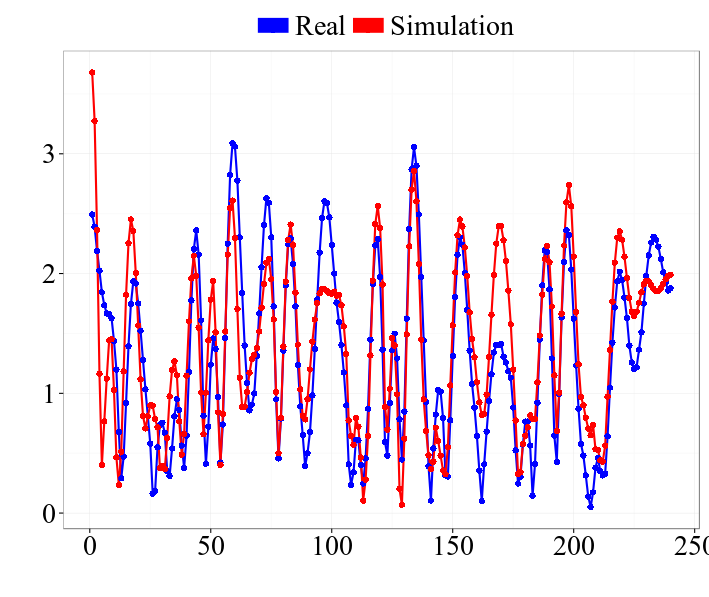}
\caption{MLP-NARX - \textit{Drives} dataset.}
\end{subfigure}
\begin{subfigure}[b]{.32\linewidth}
\includegraphics[width=\linewidth]{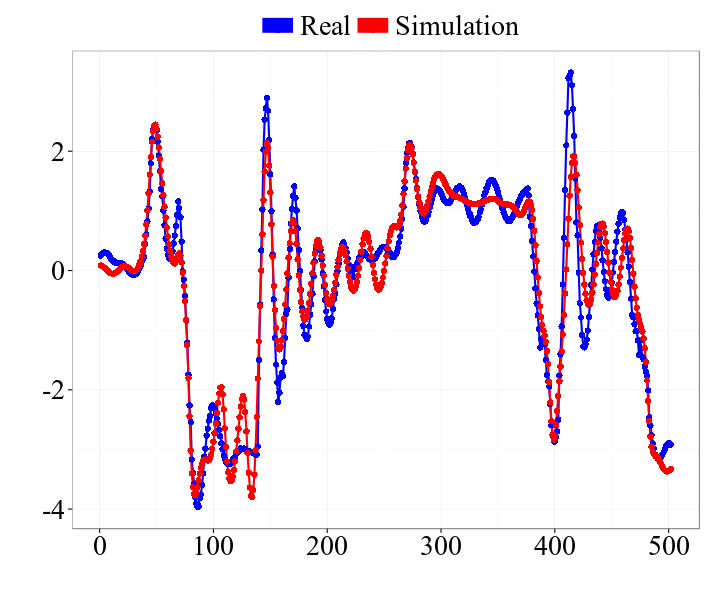}
\caption{MLP-NARX - \textit{Actuator} dataset.}
\end{subfigure}
\end{minipage}
\vskip\baselineskip
\begin{minipage}{\linewidth}
\begin{subfigure}[b]{.32\linewidth}
\includegraphics[width=\linewidth]{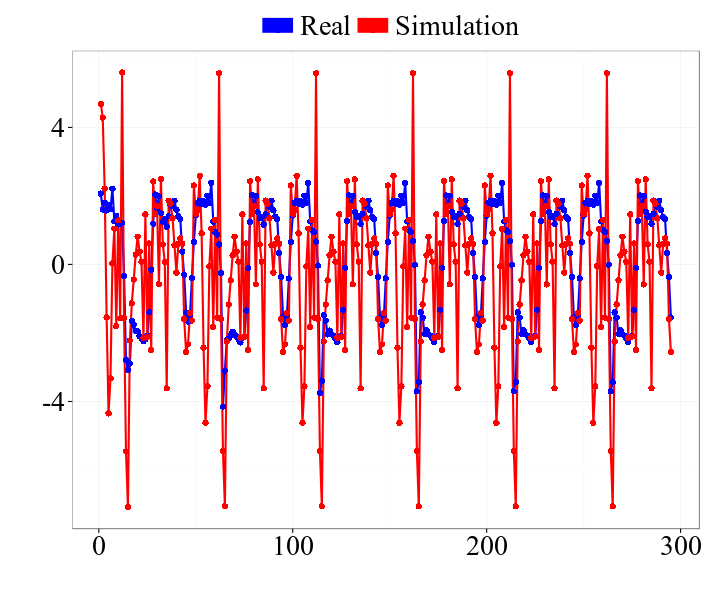}
\caption{LSTM - Artificial dataset.}
\end{subfigure}
\begin{subfigure}[b]{.32\linewidth}
\includegraphics[width=\linewidth]{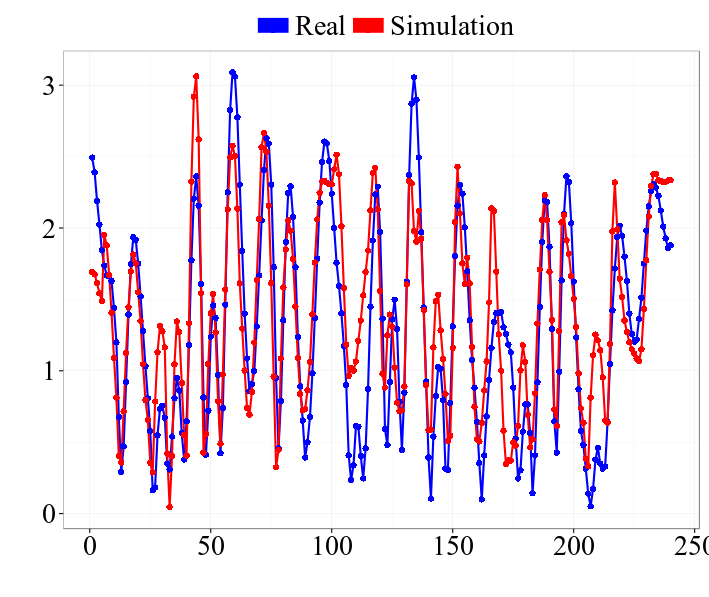}
\caption{LSTM - \textit{Drives} dataset.}
\end{subfigure}
\begin{subfigure}[b]{.32\linewidth}
\includegraphics[width=\linewidth]{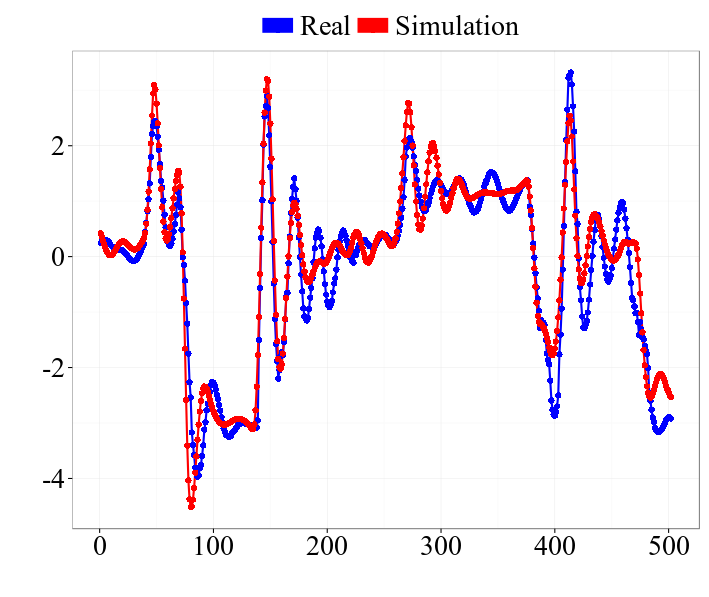}
\caption{LSTM - \textit{Actuator} dataset.}
\end{subfigure}
\end{minipage}
\vskip\baselineskip
\begin{minipage}{\linewidth}
\begin{subfigure}[b]{.32\linewidth}
\includegraphics[width=\linewidth]{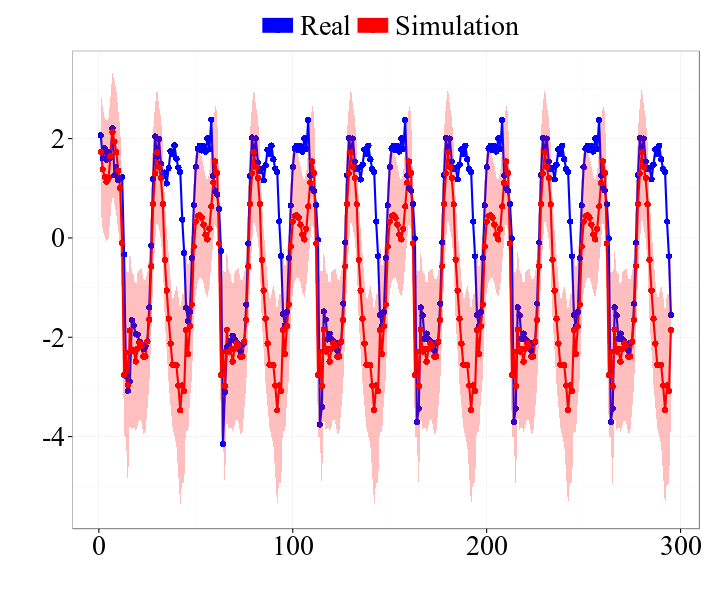}
\caption{GP-NARX - Artificial dataset.}
\end{subfigure}
\begin{subfigure}[b]{.32\linewidth}
\includegraphics[width=\linewidth]{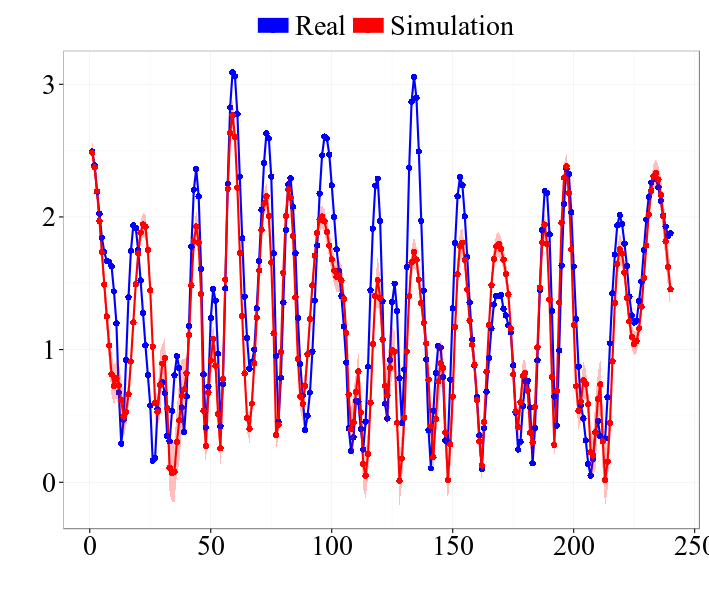}
\caption{GP-NARX - \textit{Drives} dataset.}
\end{subfigure}
\begin{subfigure}[b]{.32\linewidth}
\includegraphics[width=\linewidth]{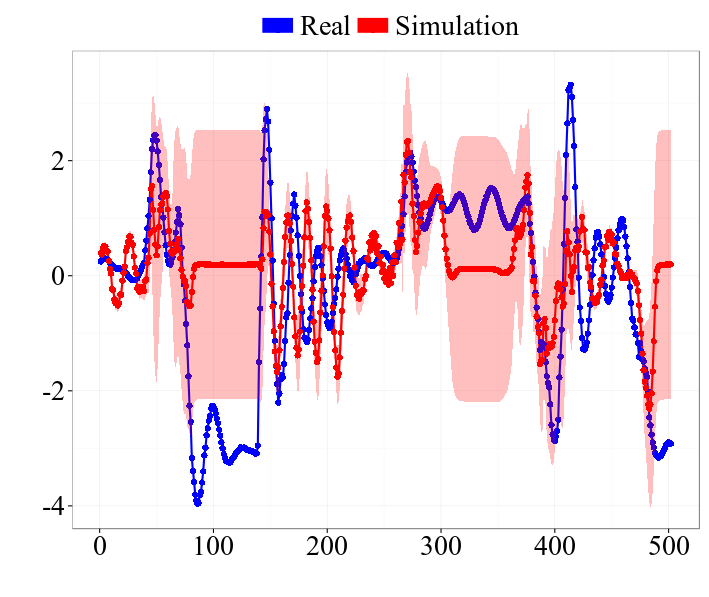}
\caption{GP-NARX - \textit{Actuator} dataset.}
\end{subfigure}
\end{minipage}
\vskip\baselineskip
\begin{minipage}{\linewidth}
\begin{subfigure}[b]{.32\linewidth}
\includegraphics[width=\linewidth]{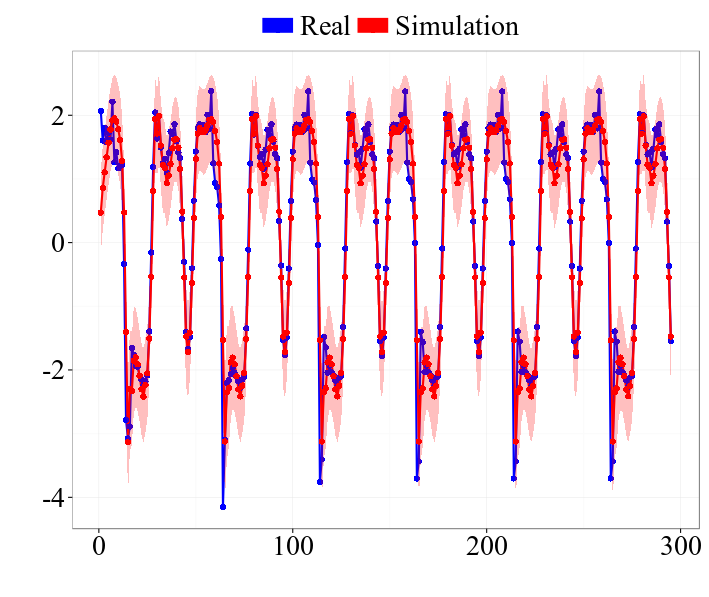}
\caption{REVARB - Artificial dataset.}
\end{subfigure}
\begin{subfigure}[b]{.32\linewidth}
\includegraphics[width=\linewidth]{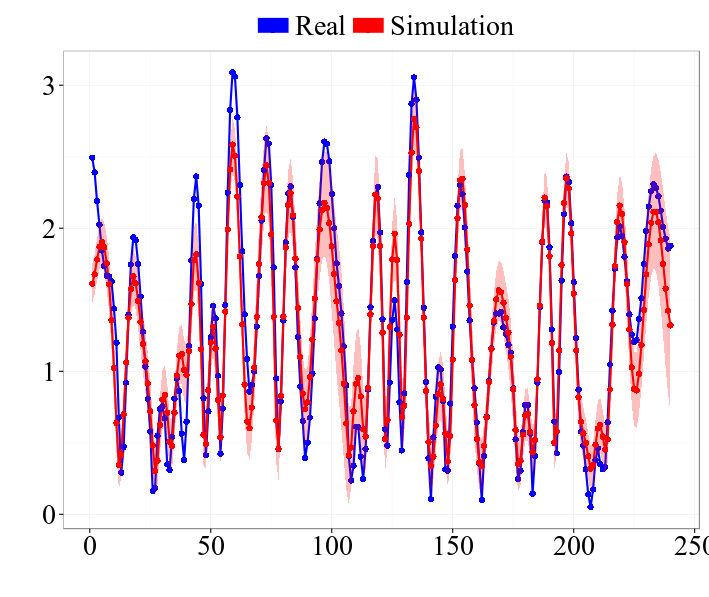}
\caption{REVARB - \textit{Drives} dataset.}
\end{subfigure}
\begin{subfigure}[b]{.32\linewidth}
\includegraphics[width=\linewidth]{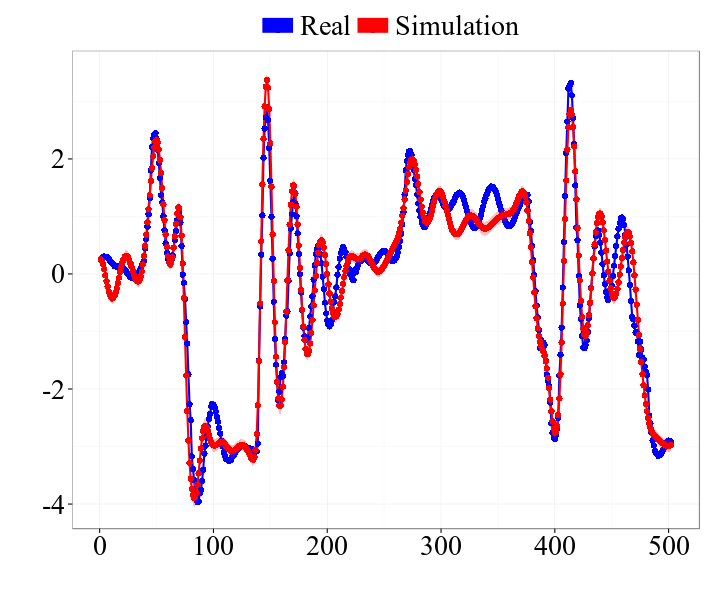}
\caption{REVARB - \textit{Actuator} dataset.}
\end{subfigure}
\end{minipage}
\caption{Free simulation on system identification test data.}
\label{FIG:RES_SIST_IDENT}
\end{figure}

The results are summarized in Tab. \ref{TAB:RES_SIST_IDENT} and the obtained simulations are illustrated in Fig. \ref{FIG:RES_SIST_IDENT}. The REVARB model was superior in all cases, with large improvements over GP-NARX. Although worse than REVARB, the MLP-NARX model presented good results, specially for the \textit{Actuator} dataset. The higher RMSE values obtained by the LSTM model is possibly related to the difficulties we have encountered when trying to optimize its architecture for this given task.

\begin{table}[!ht]
\caption{Summary of RMSE values for the free simulation results on system identification test data.}
\label{TAB:RES_SIST_IDENT}
\centering
\begin{tabular}{@{}lcccc@{}}
\toprule
& \multicolumn{1}{c}{MLP-NARX} &
\multicolumn{1}{c}{LSTM} & 
\multicolumn{1}{c}{GP-NARX} &
\multicolumn{1}{c}{REVARB} \\  
\hline
Artificial & 1.6334 & 2.2438 & 1.9245 & \textbf{0.4513} \\ 
\textit{Drive} & 0.4403 & 0.4329 & 0.4128 & \textbf{0.2491} \\ 
\textit{Actuator} & 0.4621 & 0.5170 & 1.5488 & \textbf{0.3680} \\ 
\bottomrule
\end{tabular}
\end{table}


\subsection{Human Motion Modeling}

The motion capture data from the CMU database\footnote{Available at \url{http://mocap.cs.cmu.edu/}.} was used to model walking and running motions. Training was performed with the trajectories 1 to 4 (walking) and 17 to 20 (running) from subject 35. The test set is comprised by the trajectories 5 to 8 (walking) and 21 to 24 (running) from the same subject. The original dataset contains 59 outputs, but 2 are constant, so we remove those and use the remaining 57. 

In order to perform free simulation in the test set, we include a control input given by the y coordinate of the left toes. Following the previous system identification experiments, predictions are made based only on such control input and previous predictions. We normalize the inputs and outputs with zero mean and unitary standard deviation.

We evaluate a 2 hidden layer REVARB with 200 inducing inputs, the standard GP-NARX model and a 1 hidden layer MLP with 1000 hidden units. The orders are fixed at $L = L_u = 20$. Note that the data related to both walking and running is used in the same training step. The latent autoregressive structure of REVARB allow us to train a single model for all outputs. In the case of GP-NARX, we had to train separate models for each output, since training a single model with $57 \times 20 + 20 = 1160$ dimensional regressor vector was not feasible.

The mean of the test RMSE values are summarized in Tab. \ref{TAB:RES_MOCAP}. The REVARB model obtained better results than both the other models. We emphasize that REVARB has an additional advantage over GP-NARX because its latent autoregressive structure allows the training of a single mode for all the outputs.

\begin{table}[!ht]
\caption{Summary of RMSE values for the free simulation results on human motion test data.}
\label{TAB:RES_MOCAP}
\centering
\begin{tabular}{@{}lccc@{}}
\toprule
& \multicolumn{1}{c}{MLP-NARX} &
\multicolumn{1}{c}{GP-NARX} &
\multicolumn{1}{c}{REVARB} \\  
\hline
 & 1.2141 & 0.8987 & \textbf{0.8600} \\
\bottomrule
\end{tabular}
\end{table}


\subsection{Avatar control}

\begin{figure}
        \centering
    \includegraphics[width=.8\linewidth]{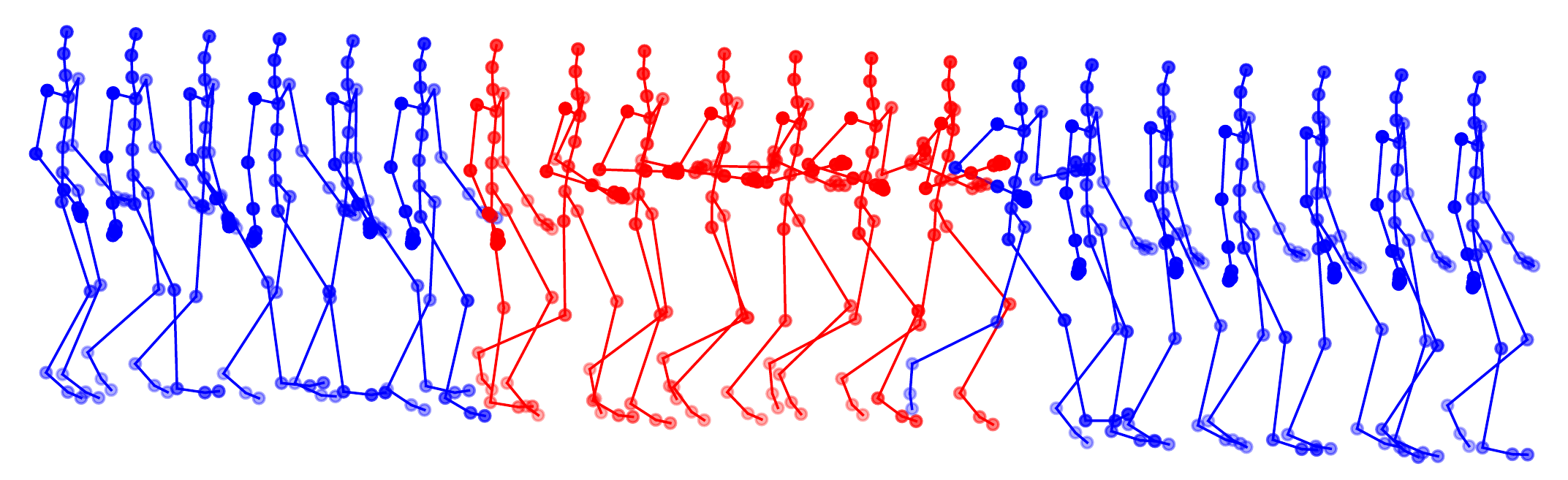}
    \caption{The generated motion with a step function signal, starting with walking (\textcolor{blue}{blue}), switching to running (\textcolor{red}{red}) and switching back to walking (\textcolor{blue}{blue}).}\label{fig:mocap_control}
\end{figure}

We demonstrate the capability of RGP by applying it to synthesize human motions with simple control signals such as the velocity. Such system ideally can be used to generate realistic human motion according to human instruction in virtual environment such as video games. We use the 5 walking and 5 running sequences from CMU motion database and take the average velocity as the control signal. We train a 1 hidden layer REVARB model with the RNN sequential recognition model (two hidden layer 500-200 units). After training, we use the model to synthesize motions with unseen control signals. Figure \ref{fig:mocap_control} shows the frames of the generated motion with a step function signal (the training sequences do not contain any switch of motions). The video of this and some more motions are available at \url{https://youtu.be/FuF-uZ83VMw}, \url{https://youtu.be/FR-oeGxV6yY}, \url{https://youtu.be/AT0HMtoPgjc}.

\section{Discussion and Further Work}
\label{SEC:CONCLUSIONS}

We defined the broad family of Recurrent Gaussian Processes models, which, similarly to RNNs, are able to learn, possibly deep, temporal representations from data. We also proposed a novel RGP model with a latent autoregressive structure where the intractabilities brought by the recurrent GP priors are tackled with a variational approximation approach, resulting in the REVARB framework. Furthermore, we extended REVARB with a sequential RNN-based recognition model that simplifies the optimization.

We applied REVARB to the tasks of nonlinear system identification and human motion modeling. The good results obtained by our model indicate that the latent autoregressive structure and our variational approach were able to better capture the dynamical behavior of the data.

In the work \cite{turner2008}, the authors present some concerns with respect to the use of mean-field approximations within a time-series context, suggesting that such approximation has a hard time propagating uncertainty through time. However, we observed in practice that our proposed REVARB framework is able to better account for uncertainty in the latent space with its autoregressive deep structure. This may be because the next layer is able to `compensate' the mean-field assumption of the previous layer, accounting for additional (temporal) correlations. Since each latent variable $x_i$ and, thus, its associated variational parameters, is present in two layers (see Eq. \ref{EQ:X_HAT}), this effect is enabled for all latent variables of the model. A similar observation is made for regular deep GPs by \cite{damianou:thesis15}.

The flexibility of GP modeling along with expressive recurrent structures is a theme for further theoretical investigations and practical applications. For instance, we intend to verify if some of the recommendations for deep modeling described by \cite{duvenaud2014} would be helpful for our RGP model.
Finally, we hope that our paper opens up new directions in the study of the parallels between RGPs and RNNs. To this end, we intend to explore the REVARB approach within longer term memory tasks and extend it with non-Gaussian likelihood distributions. 

\textbf{Acknowledgements.} The authors thank the financial support of CAPES, FUNCAP, NUTEC, CNPq (Maresia, grant 309451/2015-9, and Amalia, grant 407185/2013-5), RADIANT (EU FP7-HEALTH Project Ref 305626) and WYSIWYD (EU FP7-ICT Project Ref 612139).

{\small
\bibliography{iclr2016_conference}
\bibliographystyle{iclr2016_conference}
}

\section*{Appendix}
\subsection*{REVARB Lower Bound}

Replacing the definition of the joint distribution (Eq. \ref{EQ:JOINT}) and the factorized variational distribution $Q$ (Eq. \ref{EQ:VARIATIONAL_Q}) in the Jensen's inequality of Eq. \ref{EQ:BOUND1}, we are able to cancel the terms $p\left(f_i^{(h)} \middle| \bm{z}^{(h)}, \bm{\hat{x}}_i^{(h)}\right)$ inside the $\log$:
\begin{equation}
\begin{split}
\log p(\bm{y}) & \geq \sum_{i=L+1}^N \int_{\bm{f},\bm{x},\bm{z}} q\left(\bm{x}^{(H)}\right) q\left(\bm{z}^{(H+1)}\right) p\left(f_i^{(H+1)} \middle| \bm{z}^{(H+1)}, \bm{\hat{x}}_i^{(H)}\right) \log p\left(y_i \middle| f_i^{(H+1)}\right) \\
& + \sum_{i=L+1}^N \sum_{h=1}^{H} \int_{\bm{f},\bm{x},\bm{z}} \left( \prod_{h'=1}^{H} q\left(\bm{x}^{(h')}\right) \right) q\left(\bm{z}^{(h)}\right) p\left(f_i^{(h)} \middle| \bm{z}^{(h)}, \bm{\hat{x}}_i^{(h)}\right) \log p\left(x_i^{(h)} \middle| f_i^{(h)}\right) \\
& - \sum_{h=1}^{H+1} \int_{\bm{z}} q\left(\bm{z}^{(h)}\right) \log q\left(\bm{z}^{(h)}\right) + \sum_{h=1}^{H+1} \int_{\bm{z}} q\left(\bm{z}^{(h)}\right) \log p\left(\bm{z}^{(h)}\right) \\
& - \sum_{i=L+1}^{N} \sum_{h=1}^{H} \int_{\bm{x}} q\left(x_i^{(h)}\right) \log q\left(x_i^{(h)}\right) + \sum_{i=1}^{L} \sum_{h=1}^{H} \int_{\bm{x}} q\left(x_i^{(h)}\right) \log p\left(x_i^{(h)}\right),
\end{split}
\end{equation}
where the integrals are tractable, since all the distributions are Gaussians. The expectations with respect to $\bm{x}_i^{(h)}$ give rise to the statistics $\Psi_0^{(h)}$, $\bm{\Psi}_1^{(h)}$ and $\bm{\Psi}_2^{(h)}$, defined in Eq. \ref{EQ:PSI_STATS}.

Following similar argument of \cite{king2006}, we are able to optimally eliminate the variational parameters associated with the inducing points, $\bm{m}^{(h)}$ and $\bm{\Sigma}^{(h)}$ and get to the final form of the REVARB lower bound:
\begin{align}
\begin{split}
\log p(\bm{y}) & \geq - \frac{N-L}{2} \sum_{h=1}^{H+1} \log 2\pi\sigma_h^2 - \frac{1}{2\sigma_{H+1}^2} \left( \bm{y}^\top \bm{y} + \Psi_0^{(H+1)} - \Tr\left(\left(\bm{K}_z^{(H+1)}\right)^{-1} \bm{\Psi}_2^{(H+1)}\right) \right) \\
& + \frac{1}{2} \log\left|\bm{K}_z^{(H+1)}\right| - \log\frac{1}{2} \left| \bm{K}_z^{(H+1)} + \frac{1}{\sigma_{H+1}^2}\bm{\Psi}_2^{(H+1)} \right| \\
& + \frac{1}{2(\sigma_{H+1}^2)^2} \bm{y}^\top \bm{\Psi}_1^{(H+1)} \left( \bm{K}_z^{(H+1)} + \frac{1}{\sigma_{H+1}^2}\bm{\Psi}_2^{(H+1)} \right)^{-1} \left(\bm{\Psi}_1^{(H+1)}\right)^\top \bm{y}  \\
& + \sum_{h=1}^{H} \left\{ - \frac{1}{2\sigma_h^2} \left( \sum_{i=L+1}^{N}\lambda_i^{(h)} + \left(\bm{\mu}^{(h)}\right)^\top \bm{\mu}^{(h)} + \Psi_0^{(h)} - \Tr\left(\left(\bm{K}_z^{(h)}\right)^{-1} \bm{\Psi}_2^{(h)}\right) \right) \right. \\
& \left. + \frac{1}{2} \log\left|\bm{K}_z^{(h)}\right| - \frac{1}{2} \log\left| \bm{K}_z^{(h)} + \frac{1}{\sigma_h^2}\bm{\Psi}_2^{(h)} \right| \right. \\
& \left. + \frac{1}{2(\sigma_h^2)^2} \left(\bm{\mu}^{(h)}\right)^\top \bm{\Psi}_1^{(h)} \left( \bm{K}_z^{(h)} + \frac{1}{\sigma_h^2}\bm{\Psi}_2^{(h)} \right)^{-1} \left(\bm{\Psi}_1^{(h)}\right)^\top \bm{\mu}^{(h)} \right. \\
& \left. - \sum_{i=L+1}^N \int_{x_i^{(h)}} q\left(x_i^{(h)}\right) \log q\left(x_i^{(h)}\right) + \sum_{i=1}^{L} \int_{x_i^{(h)}} q\left(x_i^{(h)}\right) \log p\left(x_i^{(h)}\right) \right\}.
\end{split}
\end{align}

Note that the parameters of the Gaussian priors $p\left(x_i^{(h)}\right) = \mathcal{N}\left(x_i^{(h)} \middle| \mu_{0i}^{(h)}, \lambda_{0i}^{(h)} \right)$ of the initial latent variables $x_i^{(h)}|_{i=1}^L$ can be optimized along with the variational parameters and kernel hyperparameters.

\subsection*{REVARB Predictive Equations}

Predictions in the REVARB framework are done iteratively, with approximate uncertainty propagation between each layer:
\begin{align}
\mu_*^{(h)} & = \mathbb{E}\left\{ p\left(f_*^{(h)} \middle| \bm{\hat{x}}_*^{(h)}\right) \right\} = \left(\bm{B}^{(h)}\right)^\top \left(\bm{\Psi}_{1*}^{(h)}\right)^\top, \\
\begin{split}
\lambda_*^{(h)} & = \mathbb{V}\left\{ p\left(f_*^{(h)} \middle| \bm{\hat{x}}_*^{(h)}\right) \right\} = \left(\bm{B}^{(h)}\right)^\top \left( \bm{\Psi}_{2*}^{(h)} - \left(\bm{\Psi}_{1*}^{(h)}\right)^\top \bm{\Psi}_{1*}^{(h)} \right) \bm{B}^{(h)} + \Psi_{0*}^{(h)} \\
& - \Tr\left( \left( \left(\bm{K}_z^{(h)}\right)^{-1} - \left( \bm{K}_z^{(h)} + \sigma_h^{-2}\bm{\Psi}_2^{(h)} \right)^{-1}\right) \bm{\Psi}_{2*}^{(h)} \right),
\end{split}
\end{align}
where $\bm{\hat{x}}_*^{(h)}$ is defined similar to the Eq. \ref{EQ:X_HAT}, $\bm{B}^{(h)} = \sigma_h^{-2}\left( \bm{K}_z^{(h)} + \sigma_h^{-2}\bm{\Psi}_2^{(h)} \right)^{-1} \left(\bm{\Psi}_1^{(h)}\right)^\top \bm{\mu}^{(h)}$, for $1 \leq h \leq H$, and $\bm{B}^{(H+1)} = \sigma_{H+1}^{-2}\left( \bm{K}_z^{(H+1)} + \sigma_{H+1}^{-2}\bm{\Psi}_2^{(H+1)} \right)^{-1} \left(\bm{\Psi}_1^{(H+1)}\right)^\top \bm{y}$. The terms $\Psi_{0*}^{(h)}$, $\bm{\Psi}_{1*}^{(h)}$ and $\bm{\Psi}_{2*}^{(h)}$ are computed as in the Eq. \ref{EQ:PSI_STATS}, but instead of the distributions $q\left(x_i^{(h)}\right)$ we use the new Gaussian approximation $q\left(x_*^{(h)}\right) = \mathcal{N}\left(x_*^{(h)} \middle| \mu_*^{(h)}, \lambda_*^{(h)}\right)$ and replace $\bm{K}_f^{(h)}$ and $\bm{K}_{fz}^{(h)}$ respectively by $\bm{K}_*^{(h)} = k\left(\bm{\hat{x}}_*^{(h)},\bm{\hat{x}}_*^{(h)}\right)$ and $\bm{K}_{*z}^{(h)} = k\left(\bm{\hat{x}}_*^{(h)},\bm{\zeta}^{(h)}\right)$.

\end{document}